\documentclass[twoside,11pt]{article}

\usepackage{automl2020}
\usepackage{epsfig}
\usepackage{graphicx}
\usepackage{amsmath}
\usepackage{amssymb}
\usepackage[table]{xcolor}
\usepackage[ruled]{algorithm2e}
\usepackage{enumitem}
\usepackage{wrapfig}
\newcommand{\jmlrBlackBox}{\rule{1.5ex}{1.5ex}}
\newcommand{\jmlrQED}{\hfill\jmlrBlackBox}
\newtheorem{claim}{Claim}
\usepackage{cancel}
\usepackage{xr}
\usepackage{multirow}
\usepackage{booktabs}
\newcommand{\specialcell}[2][c]{%
  \begin{tabular}[#1]{@{}c@{}}#2\end{tabular}}

\SetCommentSty{mycommfont}

\SetKwInput{KwInput}{Input}                
\SetKwInput{KwOutput}{Output}

\jmlrheading{S. Dhar, U. Kurup, M. Shah}

\ShortHeadings{Stabilizing Bi-Level Hyperparameter Optimization using Moreau-Yosida Regularization}{Dhar et. al}
\firstpageno{1}

\begin{document}

\title{Stabilizing Bi-Level Hyperparameter Optimization using Moreau-Yosida Regularization}

\author{\name Sauptik Dhar \email sauptik.dhar@lge.com \\
       \addr America Research Lab, LG Electronics
       \AND
       Unmesh Kurup \email unmesh.kurup@lge.com \\
       \addr America Research Lab, LG Electronics
       \AND
	   Mohak Shah  \email mohak.shah@lge.com \\ 
	   \addr America Research Lab, LG Electronics    
       }

\maketitle

\begin{abstract}

This research proposes to use the Moreau-Yosida envelope to stabilize the convergence behavior of bi-level Hyperparameter optimization solvers, and introduces the new algorithm called Moreau-Yosida regularized Hyperparameter Optimization (MY-HPO) algorithm. Theoretical analysis on the correctness of the MY-HPO solution and initial convergence analysis is also provided. Our empirical results show significant improvement in loss values for a fixed computation budget, compared to the state-of-art bi-level HPO solvers.      
\end{abstract}

\section{Introduction}
Successful application of any machine learning (ML) algorithm heavily depends on the careful tuning of its hyper-parameters. However, hyper-parameter optimization (HPO) is a non-trivial problem and has been a topic of research for several decades. Most existing works broadly adopt one of the following two approaches, a) Black-Box optimization or, b) Direct optimization. The Black-Box optimization approach is agnostic of the underlying ML model and adopts an advanced search algorithm to select the hyper-parameters that minimizes the validation error. Popular examples include, Random search, Grid search, Bayesian Optimization based approaches \citep{hyperopt,spearmint} or other advanced exploration techniques like \citep{hyperband} etc. The advantage is that it is applicable to any machine learning problem. However, it does not utilize the structure of the underlying ML model, which can be exploited for faster solutions. The Direct Optimization utilizes the underlying ML algorithm structure by parameterizing the validation error as a function of the hyper-parameters, and directly optimizes it. Typical examples include, bi-level hyper-parameter optimization \citep{hypernetwork,STN,PedregosaHyperparameter16,franceschi2017forward,maclaurin2015gradient,franceschi2018bilevel,mehra2019penalty} and bound based analytic model selection \citep{chapelle2002choosing,dhar2019multiclass,cortes2017adanet} etc. Such approaches can provide significant computation improvements for HPO. However the limitations of these approaches include, a) the gradients of the validation error w.r.t the hyperparameters must exist and b) instability of the optimization problem under ill-conditioned settings. In this paper we address this unstable convergence behavior of the bi-level HPO approaches for ill-conditioned problems, and propose a new framework to improve it. The main contributions of this work are,
\begin{enumerate}
\item We propose a new algorithm called Moreau-Yosida regularized Hyperparameter Optimization (MY-HPO) to stabilize bi-level HPO for ill-conditioned problems. 
\item We provide theoretical analysis on the correctness of the proposed MY-HPO algorithm and also provide some initial convergence analysis.  
\item Finally, we provide extensive empirical results in support of our proposed approach.
\end{enumerate}

The paper is structured as follows. Section \ref{sect:bi_level} introduces the bi-level formulation for HPO and discusses the existing issues of the state-of-art bi-level HPO solvers \citep{hypernetwork,STN} under ill-conditioned settings. Section \ref{sect:stable_bi_level} introduces the Moreau-Yosida regularized Hyperparameter Optimization (MY-HPO) algorithm as a solution to stabilize bi-level HPO for such settings. Additional theoretical analysis of its solution's correctness and convergence behavior is also provided. Empirical results and discussions are provided in \autoref{sect:results}.   

\section{Bi-Level Hyperparameter Optimization (HPO)} \label{sect:bi_level}
The standard bi-level formulation for hyperparameter optimization (HPO) is given as,
\begin{align} \label{eq:bilevelHPO}
& \quad \quad \boldsymbol \lambda^* \in \underset{\boldsymbol \lambda}{\text{argmin}} \; L_V(\boldsymbol \lambda,\underset{\mathbf{w}}{\text{argmin}} \; L_T(\mathbf{w},\boldsymbol \lambda)) && \end{align}
\noindent Here, $L_V = $ Validation loss, $L_T =$ Training loss,
$\boldsymbol \lambda =$ Hyperparameters, \; $\mathbf{w} =$ Model parameters. A popular approach involves introducing a \textbf{best-response function} \citep{hypernetwork,STN} to rather solve,
\vspace{-0.1cm}
\begin{align} \label{eq:bilevelHPO_bestresponse}
& \quad   \boldsymbol \lambda^* \in \underset{\boldsymbol \lambda}{\text{argmin}} \; L_V(\boldsymbol \lambda, G_{\boldsymbol{\phi}}(\boldsymbol\lambda)) \quad \text{s.t.} \quad G_{\boldsymbol{\phi}}(\boldsymbol\lambda) \in \underset{\mathbf{w}}{\text{argmin}} \; L_T(\mathbf{w},\boldsymbol\lambda) &&
\end{align}
\noindent $G_{\boldsymbol{\phi}}(\boldsymbol\lambda)$ is the best-response function parameterized by the hyperparameters $\boldsymbol\lambda$. For simplicity we limit our discussion to only scalar (single) $\lambda$. Of course, it can be easily extended to multiple hyperparameters. The work on Stochastic Hyperparameter Optimization (SHO) \citep{hypernetwork} use a hypernetwork to model the best-response function as $G_{\boldsymbol{\phi}}(\lambda) = \lambda \boldsymbol{\phi}_1 + \boldsymbol{\phi}_0  = \Lambda \; \boldsymbol{\phi}$, where $ \boldsymbol{\phi} = \left[\begin{array}{c} \boldsymbol{\phi}_1 \\ \boldsymbol{\phi}_0 \end{array}  \right]$ and  $\Lambda = [\lambda \mathbf{I} | \mathbf{I}]$; with $\boldsymbol{\phi} \in \underset{\boldsymbol{\theta}}{\text{argmin}} \; L_T(\Lambda \; \boldsymbol{\theta}, \lambda)$; and adopts an alternate minimization of the training loss $L_T$ w.r.t $\boldsymbol \phi$ (hypernetwork parameters), and validation loss $\L_V$ w.r.t $\lambda$ (hyperparameters) to solve the HPO problem. Their proposed algorithm is provided in Appendix \ref{sec_SHOALG}. \cite{STN} adopts a similar alternating gradient approach but modifies the algorithm hypothesis class (by scaling and shifting the network hidden layers) and adds additional stability through adaptively tuning the degree of stochasticity (i.e. perturbation scale) for the gradient updates. In this work we adopt the SHO algorithm \ref{algSHO} as a representative for such alternating gradients approaches for solving the bi-level HPO. 

One major limitation of these alternating gradient based bilevel HPO algorithms is that, the convergence behavior heavily depends on the conditioning of the problem. Basically, for ill-conditioned settings, the step-size used for gradient updates need to be sufficiently small to ensure stability of such algorithms. This in turn leads to poor convergence rates (also see in our results \autoref{sect:results}). To alleviate this instability and ensure improved convergence we introduce our Moreau-Yosida regularized Hyperparameter Optimization (MY-HPO) algorithm. For simplicity we will assume unique solutions for \eqref{eq:bilevelHPO_bestresponse} in the rest of the paper. 

\section{Moreau-Yosida regularized HPO} \label{sect:stable_bi_level}
First we reformulate the problem in \eqref{eq:bilevelHPO_bestresponse}. Given, $\boldsymbol{\phi}^* \in \underset{\boldsymbol{\theta}}{\text{argmin}} \; L_T(\Lambda^* \; \boldsymbol{\theta}, \lambda^*)$ we solve, \vspace{-0.25cm}
\begin{flalign} \label{eqHPOADMM}
\underset{\lambda, \mathbf{w}}{\text{min}}& \quad  L_T(\mathbf{w},\lambda^*) \;+ \; L_V(\lambda,G_{\boldsymbol{\phi^*}}(\lambda))&& \\
\text{s.t.}& \quad \mathbf{w} = G_{\boldsymbol{\phi^*}}(\lambda) =  \lambda \boldsymbol{\phi}_1^* + \boldsymbol{\phi}_0^*  = \Lambda \; \boldsymbol{\phi}^* \; ; \quad \quad \text{where } \quad \boldsymbol{\phi}^* = \left[\begin{array}{c} \boldsymbol{\phi}_1^* \\ \boldsymbol{\phi}_0^* \end{array}  \right] \quad \text{and} \; \Lambda = [\lambda \mathbf{I} | \mathbf{I}] \nonumber &&
\end{flalign} 
Here, we assume that $\lambda^*$ and $\boldsymbol \phi^*$ are provided to us by an oracle satisfying $\nabla_{\boldsymbol \phi}L_T(\Lambda^* \; \boldsymbol{\phi},\lambda^*) = 0$ at solution $\Lambda^* = [\lambda^* \mathbf{I} | \mathbf{I}]$. Proposition \ref{prop_eqv} justifies solving \eqref{eqHPOADMM} in lieue of \eqref{eq:bilevelHPO_bestresponse}.

\begin{proposition} \label{prop_eqv}
For a bijective mapping $\mathbf{w} = G_{\boldsymbol{\phi^*}}(\lambda)$, the stationary points $(\mathbf{w}^*,\lambda^*)$ of \eqref{eqHPOADMM} are also stationary points of \eqref{eq:bilevelHPO_bestresponse} with $\nabla_{\mathbf{w}}L_T(\mathbf{w}^*,\lambda^*) = 0$ and $\nabla_{\lambda} L_V(\lambda^*, G_{\boldsymbol \phi^*}(\lambda^*)) = 0$.
\end{proposition}

The problem \eqref{eqHPOADMM} is a sum of two functions parameterized by different arguments connected through an equality constraint. Such formulations are frequently seen in machine learning problems and popularly solved through variants of Alternating Direction Method of Multipliers \citep{boyd2011distributed,goldstein2014fast}, Alternating Minimization Algorithm \citep{tseng1991applications} or Douglas Rachford Splitting \citep{eckstein1992douglas} etc. However, a major difference in \eqref{eqHPOADMM} is that the oracle solution $\lambda^*$ and $\boldsymbol{\phi}^* \in \underset{\boldsymbol{\theta}}{\text{argmin}} \; L_T(\Lambda^* \; \boldsymbol{\theta}, \lambda^*) \Rightarrow \nabla_{\boldsymbol \theta}L_T(\Lambda^* \; \boldsymbol{\phi}^*, \lambda^*) = 0$ is not available a priori. Hence, we cannot directly apply these existing approaches to solve \eqref{eqHPOADMM}. This leads to our new algorithm called Moreau-Yosida regularized Hyperparameter Optimization (MY-HPO) to solve \eqref{eqHPOADMM}. At iteration $k+1$ we take the steps, \\
\textbf{Step 1.} Update $\phi^{k+1}  = [\boldsymbol{\phi}_1^{k+1} | \boldsymbol{\phi}_0^{k+1}]$
\vspace{-0.5cm}
\begin{flalign} &  \mathbf{v}^{k+1} = \underset{\mathbf{v}}{\text{argmin}}\; L_T(\mathbf{v},\lambda^k) ; \quad \boldsymbol{\phi}_0^{k+1} = \overline{\mathbf{v}^{k+1}} = \sum_j v_j^{k+1}; \quad  \boldsymbol{\phi}_1^{k+1} = \frac{\mathbf{v}^{k+1} - \overline{\mathbf{v}^{k+1}}}{\lambda^k}  \label{eq_phi2} && 
\end{flalign}
\vspace{-0.7cm}
\begin{flalign} \label{eq_w_update}
    & \textbf{Step 2. } \mathbf{w}^{k+1} = \underset{\mathbf{w}}{\text{argmin}}\;  L_T(\mathbf{w},\lambda^k) + (\mathbf{u}^k)^T(\mathbf{w} - \Lambda^k \boldsymbol{\phi}^{k+1}) + \frac{\rho}{2}||\mathbf{w} - \Lambda^k \boldsymbol{\phi}^{k+1}||_2^2&&
\end{flalign}
\vspace{-0.7cm}
\begin{flalign} \label{eq_l_update}
    & \textbf{Step 3. } \mathbf{\lambda}^{k+1} = \underset{\mathbf{\lambda}}{\text{argmin}}\; L_V(\lambda, G_{\boldsymbol{\phi^{k+1}}}(\lambda)) + (\mathbf{u}^k)^T(\mathbf{w}^{k+1} - \Lambda \boldsymbol{\phi}^{k+1}) + \frac{\rho}{2}||\mathbf{w}^{k+1} - \Lambda \boldsymbol{\phi}^{k+1}||_2^2&&
\end{flalign}
\vspace{-0.7cm}
\begin{flalign} 
\text{\textbf{Step 4.} Update consensus }  &\mathbf{u}^{k+1} = \mathbf{u}^k + \rho(\mathbf{w}^{k+1} - \Lambda^{k+1} \boldsymbol{\phi}^{k+1}) && 
\end{flalign}

The complete algorithm is provided in Appendix \ref{sec_MYHPOALG}   (Algorithm \ref{algMYHPO}). Step $1$, ensures a unique solution for given $\lambda^k$ i.e. $\boldsymbol{\phi}^{k+1} = \underset{\boldsymbol{\theta}}{\text{argmin}} \; L_T(\Lambda^k \; \boldsymbol{\theta}, \lambda^k)$. In Steps $2$ and $3$ rather than taking the gradient updates of $L_T, L_V$ (as in SHO); we take the gradient of the Moreau-Yosida (MY)  regularized functions. The Moreau-Yosida (MY) regularization of a function is defined as $f_{1/\rho}(\cdot):= \underset{x}{\text{min}} ~ f(x) + \frac{\rho}{2} ||{\cdot-x}||_2 $ and serves as a smooth approximate for $f(\cdot)$. Note that \eqref{eq_w_update} and \eqref{eq_l_update} transforms to gradient updates of the MY regularized $L_T$ and $L_V$ in its scaled form \citep{boyd2011distributed}. This lends to better stability of these updates in Steps $2-3$; and is highly desirable for ill-conditioned problems. Another aspect of this algorithm is that now $\mathbf{w}$ and $\lambda$ updates are not agnostic of each other. The $\mathbf{w} - {\Lambda \phi}$ terms constrains against larger steps in a direction detrimental to either of the loss functions $L_T, L_V$. In essence these additional terms maintain consensus between the updates while minimizing $L_T$ w.r.t $\mathbf{w}$, and $L_V$ w.r.t $\lambda$ separately. Any deviation from this consensus is captured in $\mathbf{u}$-update and fed back in the next iteration. The user-defined, $\rho \geq 0$ controls the scale of these augmented terms and hence convergence of the algorithm. The proposed algorithm has the following convergence properties,

\begin{proposition} \normalfont{(Convergence Criteria)} \label{prop_stopping_criteria}
The necessary and sufficient conditions for Steps $1-4$ to solve \eqref{eqHPOADMM} are, $r^{k+1} = \mathbf{w}^{k+1} - \Lambda^{k+1} \boldsymbol{\phi}^{k+1} \rightarrow 0$ and $s^{k+1} = \rho(\Lambda^{k+1} \boldsymbol{\phi}^{k+1} - \Lambda^k \boldsymbol{\phi}^{k+1}) \rightarrow 0$.
\end{proposition} 
Further we make the following claim on the algorithm's convergence to its stationary points.
\begin{claim} \label{convergence} \normalfont{(Convergence Guarantee)}
Under the assumptions that $L_T$, $L_V$ are proper, closed and convex; with $L_T$ strongly convex and $\{ \boldsymbol \phi^k \}$ is bounded $\forall k$. The steps $1-4$ converges to a stationary point. Further at stationary point we have $r^{k+1} = 0$ and $s^{k+1} = 0$.
\end{claim}

All proofs are provided in Appendix. To maintain same per-step iteration cost as the SHO algorithm we simplify the steps 1-4 further and rather take single gradient updates in Steps 1-3. This results to the simplified MY-HPO Algorithm \ref{algMYHPO_simple} which is used throughout our experiments. For simplicity we call this simplified Algorithm \ref{algMYHPO_simple} as MY-HPO through out the paper. Additionally, to ensure descent direction we also add a backtracking (BT) of step-size scaled by a factor of $0.5$ (see chapter 4 in \citep{beck2014introduction}). The refer to the algorithm with backtracking as MY-HPO(BT) and without it i.e. constant step-size as MY-HPO (C).

\begin{algorithm}[t] 
\KwInput{\textsc{MaxIters} = Total iterations, \quad $\varepsilon_{TOL}$ = convergence error tolerance. 
$\alpha , \beta, \delta$ = step size for gradient updates.
}
\KwOutput{$\lambda,\mathbf{w}$}

\textbf{Initialize}: $\lambda \gets -1$ and $\mathbf{v} \gets \mathbf{0}, \mathbf{w} \gets \mathbf{0}, \mathbf{u} \gets \mathbf{0}$

\While{\normalfont($k+1 \leq$ \textsc{MaxIters} \normalfont) \normalfont and \normalfont( $convergence\_error \geq \varepsilon_{TOL} $ \normalfont) }{
\noindent
\vspace{-0.3cm}
\begin{flalign} 
&\mathbf{v}^{k+1} = \mathbf{v}^{k} - \alpha \nabla_{\mathbf{v}}L_T; \; \boldsymbol{\phi}_0^{k+1} = \overline{\mathbf{v}^{k+1}}; \; \boldsymbol{\phi}_1^{k+1} = \frac{\mathbf{v}^{k+1} - \overline{\mathbf{v}^{k+1}}}{\lambda} \; \text{and} \; \boldsymbol \phi^{k+1} = [\boldsymbol{\phi}_1^{k+1} | \boldsymbol{\phi}_0^{k+1}] && \nonumber
\end{flalign}
\vspace{-0.9cm}
\begin{flalign} 
    &\mathbf{w}^{k+1} =\mathbf{w}^{k}-\beta [\nabla_{\mathbf{v}}L_T + \mathbf{u}^k + \rho(\mathbf{w}^k-\Lambda^k \boldsymbol{\phi}^{k+1})] \nonumber &&
\end{flalign}
\vspace{-0.9cm}
\begin{flalign} 
&\mathbf{\lambda}^{k+1} = \mathbf{\lambda}^{k}-\delta \nabla_{\lambda} [L_V(\lambda, G_{\boldsymbol{\phi}}(\lambda)) + (\mathbf{u}^k)^T(\mathbf{w}^{k+1} - \Lambda \boldsymbol{\phi}^{k+1}) + \frac{\rho}{2}||\mathbf{w}^{k+1} - \Lambda \boldsymbol{\phi}^{k+1}||_2^2 ] && \nonumber
\end{flalign} 
\vspace{-1cm}
\begin{flalign} 
&\mathbf{u}^{k+1} = \mathbf{u}^k + \rho(\mathbf{w}^{k+1} - \Lambda^{k+1} \boldsymbol{\phi}^{k+1}) && \nonumber
\end{flalign}
\vspace{-0.8cm}
}
\caption{Simplified MY-HPO Algorithm \label{algMYHPO_simple}}
\end{algorithm}  


\section{Results and Summary} \label{sect:results}
\subsection{Experimental Settings}
We provide analysis for both regression and classification problems using the loss functions, 
\begin{itemize}[nosep,leftmargin=*]
\item[--] Least Square: $L_V = \frac{1}{2N_V} \sum_{\mathbf{x}_i \in \mathcal{V}} (y_i - \mathbf{w}^T\mathbf{x}_i)^2$, $L_T = \frac{1}{2N_T} \sum_{\mathbf{x}_i \in \mathcal{T}} (y_i - \mathbf{w}^T\mathbf{x}_i)^2 + e^\lambda ||\mathbf{w}||_2^2$
\item[--] Logistic: $L_V = \frac{1}{N_V} \sum_{\mathbf{x}_i \in \mathcal{V}} \text{log}(1+e^{-y_i\mathbf{w}^T\mathbf{x}_i})$, $L_T = \frac{1}{N_T} \sum_{\mathbf{x}_i \in \mathcal{T}} \text{log}(1+e^{-y_i\mathbf{w}^T\mathbf{x}_i}) + e^\lambda ||\mathbf{w}||_2^2$
\end{itemize}
Here, $\mathbf{x}_i \in \Re^d$, and $\mathbf{y} \in \Re$ (regression) or $\{-1,1\}$ (classification), $d$ = dimension of the problem, $\mathcal{T}$ = Training data, $N_T$ = number of training samples and $\mathcal{V}$ = Validation Data, $N_V$ = number of validation samples. The experimental details are provided in Table \ref{tab_data}. Here, for regression, we use the MNIST data to regress on the digit labels `0' - `9' \citep{vmpg} and the \textsc{Cookie} to predict the percentage of `fat' w.r.t the `flour' content using near infrared region (NIR) values \citep{osborne1984application}. For classification, we classify between digits `0' vs. `1' for MNIST, and the traffic signs `30' vs. `80' for GTSRB.

\begin{table}
\centering
\caption{Datasets and Experimental Settings}. 
\label{tab_data}
\tabcolsep=0.1cm
\begin{scriptsize}
\begin{sc}
\begin{tabular}{|l|ccc|}  
\hline 
\quad \quad \quad \quad \quad \quad Data Sets & \specialcell{Training/Validation size} & \specialcell{Test size} & \specialcell{Dimension ($d$)} \\ 
\hline
\specialcell{\textbf{COOKIE} \citep{osborne1984application}} & \specialcell{$34/17\; (50\%/25\%)$} & \specialcell{$19 \;(25\%)$} & 699 (NIR) \\ 
\specialcell{\textbf{MNIST} Regression \citep{lecun1998mnist}}  & \specialcell{$1000$ ($100$ per digit)} &  \specialcell{$5000$ ($500$ per digit)} & \specialcell{$784$ (Pixel)} \\
\specialcell{\textbf{MNIST} Classification \citep{lecun1998mnist}}  & \specialcell{$500$ ($250$ per class)} &  \specialcell{$1000$ ($500$ per class)} & \specialcell{$784$ (Pixel)} \\
\specialcell{\textbf{GTSRB} \citep{stall12}} & \specialcell{$1000$ ($500$ per class)}  & \specialcell{$1000$ ($500$ per class)} & $1568$ (HOG)  \\
\hline
\end{tabular}
\end{sc}
\end{scriptsize}
\end{table}

\subsection{Results}
Table \ref{tab_results} provides the average ($\pm$ standard deviation) of the loss values on training,validation and a separate test data, over 10 runs of the experiments. Here, for each experiment we partition the data sets in the same proportion as shown in Table \ref{tab_data}. For regression we scale the output's loss with the variance of y $-$ values. This is a standard normalization technique which illustrates the proportion of un-explained variance by the estimated model. For the SHO and MY-HPO algorithms we only report the performance of the models using the best performing step-sizes. A detailed ablation study using different step-sizes for the algorithms is provided in Appendix \ref{sec_params}. In addition we also provide the results for popular black-box algorithms publicly available through the Auptimizer toolbox \citep{liu2019auptimizer}. Here for all the algorithms we maintain a fixed budget of gradient computations. Additional details on the various algorithm's parameter settings are provided in Appendix \ref{sec_params}.  

\begin{figure*}
\centering
\includegraphics[height=5cm]{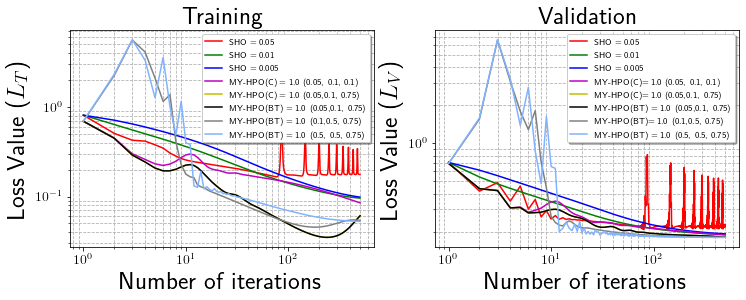}
\caption{Convergence behavior of SHO vs. MY-HPO for different step-sizes for GTSRB data using Logistic loss. For MY-HPO we show the $\rho (\alpha,\beta,\delta)$ values.}\label{fig_GTSTB_all}
\vspace{-0.2cm}
\end{figure*}

Table \ref{tab_results} shows that for both the classification and regression problems MY-HPO algorithm significantly outperforms all the baseline algorithms. The dominant convergence behavior of the bi-level formulations (i.e. MY-HPO and SHO) compared to black-box approaches is well known from previous studies \citep{hypernetwork,STN}. This is also seen in our results. Of course, increasing the number of gradient computations allows the black-box approaches achieve similar loss values (see Appendix \ref{sec_params}). However, a non-trivial observation is that MY-HPO significantly outperforms the SHO algorithm. To further analyze this improved convergence behavior of the MY-HPO algorithm; we provide the convergence curves of these algorithms for Logistic loss using GTSRB data in Fig \ref{fig_GTSTB_all}. Here, MY-HPO (BT) involves back-tracking to ensure descent direction; whereas MY-HPO (C) uses constant step updates. Fig \ref{fig_GTSTB_all} shows that the SHO algorithm obtains best results with the step-size fixed to $\alpha = 0.01$. Increasing it further to $\alpha = 0.05$ destabilizes the updates and results to sub-optimal SHO solution. On the other hand, using the Moreau Yosida regularized updates we can accomodate for larger step-sizes and hence achieve better convergence rates. Additional improvement can be expected by using back-tracking as it ensures descent directions which adds to the stability of the algorithm.  This behavior is persistently seen for both the classification and regression problems for all the data sets (see Appendix \ref{sec_params}).

\subsection{Summary}
In summary our results confirm the effectiveness of Moreau-Yosida (MY) regularized updates for bi-level HPO under (ill-conditioned) limited data settings. Here rather than taking alternating gradient updates as in SHO \citep{hypernetwork} or STN \citep{STN}; we propose a modified algorithm MY-HPO by taking gradient updates on the Moreau-Yosida envelope and maintaining a consensus variable. The proposed MY-HPO algorithm provides added stability and enables us to use larger step sizes which inturn leads to better convergence rates. Under a fixed computation budget MY-HPO significantly outperforms the  SHO algorithm (a popular representative for alternating gradient based bi-level HPO solvers); and the widely used black-box HPO routines. Owing to space constraints details regarding the algorithm parameters for reproducing the experimental results are provided in Appendix \ref{sec_params}.

\begin{table}
\centering
\caption{Loss values of different HPO algorithms using fixed computation budget.} 
\label{tab_results}
\tabcolsep=0.08cm
\begin{scriptsize}
\begin{sc}
\begin{tabular}{|c|ccccccc|}  
\hline  
Data &SHO& \specialcell{MyHPO (C)}& \specialcell{MyHPO (BT)} & Random & Grid  & HyperOpt & Spearmint \\
\hline
\multicolumn{8}{c}{\textbf{Regression Problems}} \\
\hline
\textbf{Cookie} & \multicolumn{6}{c}{\text{No. of gradient computations = 5000 }}&\\
Train ($\times 10^{-2}$) &$71.5 \pm 66.9$&$7.7 \pm 2.7$& $\mathbf{5.4 \pm 2.5}$ & $89.1 \pm 10.3$ &$6.7 \pm 2.2$ & $58.1 \pm 7.9$  & $6.7 \pm 2.2$  \\ 
Val. ($\times 10^{-2}$) &$63.1 \pm 72.8$&$17.6 \pm 13.4$& $\mathbf{6.2 \pm 4.8}$& $93.4 \pm 26.4$ &$16.7 \pm 12.9$ & $49.7 \pm 19.5$  & $16.7 \pm 12.9$ \\ 
Test ($\times 10^{-2}$) & $75.3 \pm 98.7$ & $17.4 \pm 15.8$& $\mathbf{6.9 \pm 7.3}$ &$85.6 \pm 28.9$ &$16.1 \pm 14.7$ & $47.2 \pm 23.5$  & $16.1 \pm 14.7$\\ [.5mm]\hline
\textbf{MNIST} & \multicolumn{6}{c}{\text{No. of gradient computations = 6000 }} &\\
Train($\times 10^{-2}$) &$26.7 \pm 2.7$  & $\mathbf{22.6 \pm 0.6}$ & $\mathbf{22.6 \pm 0.6}$ & $22.3 \pm 0.6$  & $17.4 \pm 0.5$ &  $22.8 \pm 5.7$  & $17.4 \pm 0.5$\\ 
Val. ($\times 10^{-2}$) & $24.6 \pm 1.4$   & $\mathbf{23.7 \pm 1.0}$  & $\mathbf{23.7 \pm 1.0}$ & $25.6 \pm 2.8$ & $26.3 \pm 2.4$ & $25.0 \pm 2.5$  & $26.3 \pm 2.5$\\ 
Test ($\times 10^{-2}$) & $23.4 \pm 1.1$ & $\mathbf{22.3 \pm 0.55}$ & $\mathbf{22.3 \pm 0.55}$ & $24.6 \pm 2.8$  &  $24.8 \pm 2.2$& $23.8 \pm 2.1$  & $24.8 \pm 2.2$ \\ \hline
\multicolumn{8}{c}{\textbf{Classification Problems}} \\
\hline
\textbf{MNIST} &\multicolumn{6}{c}{\text{No. of gradient computations = 1000 }} &\\
Train ($\times 10^{-2}$) &$5.58 \pm 2.1$&$2.94 \pm 1.6$& $\mathbf{2.87 \pm 1.6}$ & $35.7 \pm 71.4$ & $0.3 \pm 0.1$&  $5.18 \pm 0.35$ & $0.3 \pm 0.1$\\ 
Val. ($\times 10^{-2}$) &$4.9 \pm 1.96$& $4.4 \pm 1.89$ &  $\mathbf{4.3 \pm 1.91}$ & $19.1 \pm 37.8$ &$5.31 \pm 2.7$ &  $4.9 \pm 1.46$  & $5.31 \pm 2.7$\\ 
Test ($\times 10^{-2}$) & $5.32 \pm 1.6$ & $5.1 \pm 1.84$ & $\mathbf{5.0 \pm 1.79}$ & $19.4 \pm 38.6$& $6.22 \pm 2.8$ & $5.19 \pm 1.44$ & $6.22 \pm 2.8$ \\ [.5mm]\hline
\textbf{GTSRB} & \multicolumn{6}{c}{\text{No. of gradient computations = 1000}} &\\
Train ($\times 10^{-2}$) & $6.83 \pm 1.04$ & $3.26 \pm 2.5$ & $\mathbf{4.31 \pm 1.2}$ & $47.48 \pm 117.3$ & $0.09 \pm 0.01$  &  $11.9 \pm 30.5$  & $0.09 \pm 0.01$ \\ 
Val. ($\times 10^{-2}$) & $14.35 \pm 2.34$ & $14.35 \pm 2.38$ & $\mathbf{13.98 \pm 2.3}$ & $39.85 \pm 75.4$ &  $23.5 \pm 6.6$  &  $18.9 \pm 11.7$  & $23.5 \pm 6.6$ \\ 
Test ($\times 10^{-2}$) & $14.04 \pm 1.82$ & $13.8 \pm 2.25$ & $\mathbf{13.59 \pm 1.9}$ & $38.6 \pm 70.6$ & $21.8 \pm 5.2$  & $17.6 \pm 8.5$ &$21.8 \pm 5.2$ \\ \hline
\end{tabular}
\end{sc}
\end{scriptsize}
\end{table}

\vskip 0.2in
\bibliography{bi_level_HPO}


\newpage

\appendix
\section{Proofs}
\subsection{Proof of Proposition \ref{prop_eqv}}

The proof follows from analyzing the KKT equations. Note that the lagrangian of the problem \ref{eq:bilevelHPO_bestresponse} is,
\begin{flalign} 
& \mathcal{L}(\mathbf{w}, \lambda, \mathbf{u}) = L_T(\mathbf{w},\lambda^*)+ L_V(\lambda,G_{\boldsymbol{\phi}^*}(\lambda)) + \mathbf{u}^T(\mathbf{w}-G_{\phi^*}(\lambda))&& 
\end{flalign}

\noindent \underline{KKT system},
\begin{flalign}
 \nabla_{\mathbf{w}} L_T(\mathbf{w},\lambda^*) + \mathbf{u} &=& 0 && \label{eq_w}\\
 \nabla_{\lambda}L_V(\lambda,G_{\phi^*}(\lambda)) - \mathbf{u}^T \nabla_{\lambda} G_{\phi^*}(\lambda) &=& 0 && \label{eq_v}\\
 \mathbf{w} - G_{\phi^*}(\lambda) &=& 0 && \label{eq_wG}\\
 \nabla_{\phi} L_T(G_{\phi^*}(\lambda), \lambda) &=& 0 \label{eq_lamL}
\end{flalign} 
At the solution of the KKT System $(\mathbf{w}^*,\lambda^*,\mathbf{u}^*)$
for a bijective mapping $\mathbf{w}^* = G_{\phi^*}(\lambda^*) = \Lambda^*\boldsymbol{\phi}^*$ we have eq. \eqref{eq_lamL} $\Rightarrow \nabla_{\mathbf{w}} L_T(\mathbf{w}^*,\lambda^*) = 0$.  This in turn indicates, $\mathbf{u}^* = 0 \Rightarrow \nabla_{\lambda} L_V(\lambda^*, G_{\phi^*}(\lambda^*)) = 0$. Hence at solution, $\nabla_{\mathbf{w}} L_T(\mathbf{w}^*,\lambda^*) = 0$ and $\nabla_{\lambda} L_V(\lambda^*, G_{\phi^*}(\lambda^*)) = 0$ 

\jmlrQED

\subsection{Proof of Proposition \ref{prop_stopping_criteria}}
Assume the conditions $r^{k+1} = 0$ and $s^{k+1} = 0$, are satisfied at iteration $k+1$. Then the steps $1-4$ gives,
\begin{flalign} 
& \nabla_{\mathbf{w}}L_T(\mathbf{w}^{k+1},\lambda^k) + \mathbf{u}^k + \rho (\mathbf{w}^{k+1} - \Lambda^k \boldsymbol{\phi}^{k+1}) = 0 && \label{eq_ADMM1a} \\ 
\Rightarrow & \nabla_{\mathbf{w}}L_T(\mathbf{w}^{k+1},\lambda^k) + \mathbf{u}^{k+1} + \rho (\Lambda^{k+1} \boldsymbol{\phi}^{k+1} - \Lambda^{k} \boldsymbol{\phi}^{k+1}) = 0 &&  \label{eq_ADMM1b} \\
& \nabla_{\lambda} L_V(\lambda^{k+1},\Lambda^{k+1} \boldsymbol{\phi}^{k+1}) - (\mathbf{u}^k)^T \boldsymbol{\phi}_1^{k+1} - \rho (\boldsymbol{\phi}_1^{k+1})^T(\mathbf{w}^{k+1} - \Lambda^{k+1} \boldsymbol{\phi}^{k+1})  = 0 && \label{eq_ADMM1c} \\
& \mathbf{u}^{k+1} = \mathbf{u}^k + \rho(\mathbf{w}^{k+1} - \Lambda^{k+1} \boldsymbol{\phi}^{k+1}) && \label{eq_ADMM1d}
\end{flalign}
Under the conditions $r^{k+1} = 0, s^{k+1} = 0$ the updates in eq. \eqref{eq_ADMM1a} - \eqref{eq_ADMM1d} becomes,
\begin{flalign} \label{eq_ADMM2}
& \nabla_{\mathbf{w}}L_T(\mathbf{w}^{k+1},\lambda^k) + \mathbf{u}^{k+1} = 0 && \nonumber \\
& \nabla_{\lambda} L_V(\lambda^{k+1},\Lambda^{k+1} \boldsymbol{\phi}^{k+1}) - (\mathbf{u}^{k+1})^T \boldsymbol{\phi}_1^{k+1} = 0 && \\
& \mathbf{u}^{k+1} = \mathbf{u}^k && \nonumber
\end{flalign}

\noindent In addition at iteration $k+1$ we also have from eq. \eqref{eq_phi2} a unique solution for \[(\boldsymbol{\phi}_1^{k+1}, \boldsymbol{\phi}_0^{k+1}) = \underset{\boldsymbol{\theta}_1,\boldsymbol{\theta}_0}{\text{argmin}} \; L_T(\lambda^k \boldsymbol{\theta}_1 + \boldsymbol{\theta}_0,\lambda^k)\] 
Also we have from our assumptions, 
\begin{flalign}
r^{k+1} = 0 \Rightarrow \mathbf{w}^{k+1} = \lambda^k \boldsymbol{\phi}_1^{k+1} + \boldsymbol{\phi}_0^{k+1} \Rightarrow \nabla_{\mathbf{w}} L_T(\mathbf{w}^{k+1}, \lambda^{k}) = 0 &&  \label{eq_stationary1}\\
s^{k+1} = 0 \Rightarrow \lambda^k \boldsymbol{\phi}_1^{k+1} = \lambda^{k+1} \boldsymbol{\phi}_1^{k+1} \Rightarrow \nabla_{\mathbf{w}} L_T(\mathbf{w}^{k+1}, \lambda^{k+1}) = 0 \label{eq_stationary2}&&
\end{flalign}


\noindent Using e.q. \eqref{eq_stationary1} and \eqref{eq_stationary2} into \eqref{eq_ADMM2} gives,
\begin{flalign}
& \nabla_{\mathbf{w}}\; L_T(\mathbf{w}^{k+1},\lambda^{k+1}) = 0  \quad (\Rightarrow \mathbf{u}^{k+1} = 0) && \nonumber \\
& \nabla_{\lambda} L_V(\lambda^{k+1},\Lambda^{k+1} \boldsymbol{\phi}^{k+1}) = 0 \quad (\because \mathbf{u}^{k+1} = 0)\nonumber &&\\
&\Rightarrow \nabla_{\lambda} L_V(\lambda^{k+1},\mathbf{w}^{k+1}) = 0 \quad (\text{for bijective } \Lambda\boldsymbol{\phi} \rightarrow \mathbf{w}) && \nonumber
\end{flalign} 

\noindent Next for the necessary part if $r^{k+1} \neq 0$, the primary constraint in \eqref{eqHPOADMM} is not satisfied. Hence $r^{k+1} = 0$ is a necessary condition. To establish the necessary condition $s^{k+1} = 0$, consider $\nabla_{\lambda} L_V(\lambda^{k+1},\mathbf{w}^{k+1}) = 0$ and $\nabla_{\mathbf{w}}\; L_T(\mathbf{w}^{k+1},\lambda^{k+1}) = 0$. \\

\noindent Now, $r^{k+1} = 0 \Rightarrow \mathbf{w}^{k+1} = \boldsymbol{\phi}^{k+1} \boldsymbol{\Lambda}^{k+1} \Rightarrow \nabla_{\boldsymbol{\phi}} \; L_T(\boldsymbol{\phi}^{k+1} \boldsymbol{\Lambda}^{k+1},\lambda^{k+1}) = 0$ (from above assumption). Finally, Step 1 of Algorithm \ref{algMYHPO} ensures, $\nabla_{\boldsymbol{\phi}} \; L_T(\boldsymbol{\phi}^{k+1} \boldsymbol{\Lambda}^{k},\lambda^k) = 0$. Hence $\boldsymbol{\phi}^{k+1} \boldsymbol{\Lambda}^{k} = \boldsymbol{\phi}^{k+1} \boldsymbol{\Lambda}^{k+1} \Rightarrow \mathbf{s}^{k+1} = 0$. \jmlrQED

\subsection{Proof for Claim \ref{convergence}}
For the first part of the proof observe that the steps $2-4$ are exactly the ADMM updates for a given $\boldsymbol{\phi}^{k+1}$. This allows us to re-utilze the Proposition 4 in \citep{giselsson2016linear} and claim that the operator equivalent to the steps $2-4$ is contractive. The exact form of this new operator and the equivalent convergence rates will be analyzed in a longer version of this work.

\noindent For the second part observe that the state of the system is determined by the variables $(\lambda^k,\mathbf{u}^k) \rightarrow (\lambda^{k+1},\mathbf{u}^{k+1})$. At stationary points the states remain the same. This gives,  
\begin{flalign}
\text{From \eqref{eq_ADMM1d},} & \quad \mathbf{w}^{k+1} -\Lambda^{k+1}\boldsymbol{\phi}^{k+1} = 0 \Rightarrow r^{k+1} = 0 \nonumber && \end{flalign}
\begin{flalign}
&\text{Further, } \rho (\Lambda^{k+1} \boldsymbol{\phi}^{k+1} - \Lambda^{k} \boldsymbol{\phi}^{k+1}) = 0 \quad \quad (\because \Lambda^k = \Lambda^{k+1} \text{and } \phi \text{ is bounded })\nonumber &&
\end{flalign} \jmlrQED

\section{Stochastic Hyperparameter Optimization using Hypernetworks (SHO)}  \label{sec_SHOALG}
There are several versions (global vs. local) of the SHO algorithm introduced in \citep{hypernetwork}. The representative local version of the SHO algorithm is provided below in Algorithm \ref{algSHO}.

\begin{algorithm} 
\KwInput{\vspace{-0.5cm}
\begin{flalign}
&\alpha &=& \text{learning rate of training loss gradient update w.r.t model parameters.}&& \nonumber \\
&\beta &=& \text{learning rate of validation loss gradient update w.r.t hyperparameters.}&& \nonumber 
\end{flalign}
}  \vspace{-0.25cm}
\KwOutput{$\lambda, \mathbf{G}_{\boldsymbol \phi}(\lambda)$}
\KwData{$\mathcal{T}$ = Training Data , $\mathcal{V}$ = Validation Data}
\textbf{Initialize}: $\boldsymbol\phi , \lambda$ and, \textbf{define}: $\mathbf{w} = G_{\boldsymbol{\phi}}(\lambda) = \Lambda \; \boldsymbol{\phi}$ \;
\While{\normalfont $not\_converged$ }{
    $\hat{ \lambda}\sim P(\hat{ \lambda}| \lambda^{t-1})$ \tcp{typically modeled as Normal Distribution} 
    $\boldsymbol \phi^t \gets \boldsymbol \phi^{t-1} - \alpha \nabla_{G_{\boldsymbol \phi}} L_T(G_{\boldsymbol \phi} ( \hat{\lambda}) ,  \hat{\lambda}, \mathbf{x} \in \mathcal{T}) \cdot \underset{ = \hat{\boldsymbol \Lambda}}{\nabla_{\boldsymbol \phi} G_{\boldsymbol \phi}(\hat{\lambda)}}$ \; 
    $ \lambda^{t} \gets  \lambda^{t-1} - \beta \underset{ = \boldsymbol \phi^T}{\nabla_{ \lambda} G_{\boldsymbol \phi}(\lambda)} \cdot  \nabla_{G_{\boldsymbol \phi}} L_V(G_{\boldsymbol \phi}( \lambda),\mathbf{x} \in \mathcal{V})$ 
    \vspace{-0.25cm}
  }
 \caption{SHO Algorithm (Local) \label{algSHO}}
\end{algorithm} 

\newpage
\section{Moreau Yosida regularized (MY)-HPO Algorithm} \label{sec_MYHPOALG}

\begin{algorithm}[H] 
\KwInput{\textsc{MaxIters} = Total iterations, \quad $\varepsilon_{TOL}$ = convergence error tolerance.
}
\KwOutput{$\lambda,\mathbf{w}$}
\KwData{$\mathcal{T}$ = Training Data , $\mathcal{V}$ = Validation Data}
\textbf{Initialize}: $\lambda \gets -1$ and $\mathbf{u} \gets \mathbf{0}$

\While{\normalfont($k+1 \leq$ \textsc{MaxIters} \normalfont) \normalfont and \normalfont( $convergence\_error \geq \varepsilon_{TOL} $ \normalfont) }{
\textbf{Step 1.} Update $\phi$
\begin{flalign} \mathbf{v}^{k+1} = \underset{\mathbf{v}}{\text{argmin}}\; L_T(\mathbf{v},\lambda^k)  && \nonumber
\end{flalign}
\vspace{-1.1cm}
\begin{flalign} \boldsymbol{\phi}_0^{k+1} = \overline{\mathbf{v}^{k+1}} \quad ; \quad  \boldsymbol{\phi}_1^{k+1} = \frac{\mathbf{v}^{k+1} - \overline{\mathbf{v}^{k+1}}}{\lambda^k} \quad \text{and} \quad \boldsymbol \phi^{k+1} = [\boldsymbol{\phi}_1^{k+1} | \boldsymbol{\phi}_0^{k+1}] && \nonumber
\end{flalign}

\textbf{Step 2.} Update $\mathbf{w}$
\vspace{-0.1cm}
\begin{flalign} 
    &\mathbf{w}^{k+1} = \underset{\mathbf{w}}{\text{argmin}}\;  L_T(\mathbf{w},\lambda^k) + (\mathbf{u}^k)^T(\mathbf{w} - \Lambda^k \boldsymbol{\phi}^{k+1}) + \frac{\rho}{2}||\mathbf{w} - \Lambda^k \boldsymbol{\phi}^{k+1}||_2^2&& \nonumber
\end{flalign}

\textbf{Step 3.} Update $\lambda$
\vspace{-0.1cm}
\begin{flalign} 
    &\mathbf{\lambda}^{k+1} = \underset{\mathbf{\lambda}}{\text{argmin}}\; L_V(\lambda, G_{\boldsymbol{\phi}^{k+1}}(\lambda)) + (\mathbf{u}^k)^T(\mathbf{w}^{k+1} - \Lambda \boldsymbol{\phi}^{k+1}) + \frac{\rho}{2}||\mathbf{w}^{k+1} - \Lambda \boldsymbol{\phi}^{k+1}||_2^2&& \nonumber
\end{flalign} 
\vspace{-1cm}
\begin{flalign} 
\text{\textbf{Step 4.} Update consensus }  &\mathbf{u}^{k+1} = \mathbf{u}^k + \rho(\mathbf{w}^{k+1} - \Lambda^{k+1} \boldsymbol{\phi}^{k+1}) && \nonumber
\end{flalign}
}
\caption{MY-HPO algorithm \label{algMYHPO}}
\end{algorithm} 
\noindent \textbf{Connection with ADMM} : The MY-HPO updates are closely connected to the Alternating Direction Method of Multipliers (ADMM) algorithm \citep{boyd2011distributed}. In fact, for a given $\boldsymbol \phi^k$ the steps $2 - 4$ are exactly the ADMM updates. However, the algorithm is fundamentally different from ADMM. For example, unlike ADMM, interchanging the steps $2$ and $3$ completely changes the stationary points. Moreover, the updates do not transform to the Douglas Rachford splitting operator as traditionally seen for ADMM \citep{giselsson2016linear}. Still, the similarities of MY-HPO with ADMM enables us to modify previous convergence analyses in \citep{giselsson2016linear} to obtain Proposition \ref{prop_stopping_criteria} and Claim \ref{convergence}. 
\vspace{0.5cm}

\noindent \textbf{Relaxation to simplified MY-HPO algorithm \ref{algMYHPO_simple}} : Algorithm \ref{algMYHPO_simple} simplifies the above algorithm \ref{algMYHPO} by taking one gradient update rather than solving the minimization problems above. This reduces the per-step iteration cost of the algorithm \ref{algMYHPO_simple} to be the same as the SHO algorithm in \ref{algSHO}. Now, both the Algorithms \ref{algMYHPO_simple} and \ref{algSHO}  incurs 2-gradient steps per outer iteration. This simplification however deteriorates the convergence rate of the algorithm. We provide the convergence comparison between the Algorithms \ref{algMYHPO_simple} vs. \ref{algMYHPO} in Table \ref{tab_results_closed}.
  
\begin{table}
\centering
\caption{Performance of MY-HPO Algorithm \ref{algMYHPO} for the regression problems in Table \ref{tab_results}}
\label{tab_results_closed}
\begin{sc}
\begin{tabular}{|c|ccccc|}  
\hline  
Data & Train Loss & Val. Loss & Test Loss & No. of Iterations ($k$) & MaxIters\\
\hline
\textbf{Cookie} & $0.31 \pm 0.20$ & $0.22 \pm 0.21$ & $0.26 \pm 0.25$& $80 \pm 18.68 $ & $100$\\[.5mm]  
\textbf{MNIST} & $8.85 \pm 0.35 $ & $9.85 \pm 0.47 $ &$9.32 \pm 0.35 $& $364 \pm 124.3 $&$500$\\
\hline
\end{tabular}
\end{sc}
\end{table}

\section{Algorithm Parameters and Additional Results} \label{sec_params}

\noindent For all the black-box approaches we match the number of outer-iterations of the bi-level formulations i.e. $n_T$ to train the model for any given parameter. Further, we train the models using gradient updates with step size $(\alpha_{train})$. This parameter is set specific to each problems reported below, and helps us achieve similar training gradient as the bi-level formulations. In addition, we set the following parameters using the Auptimizer tool-box for each of the following algorithms. 
\vspace{0.2cm}

\noindent \underline{Random Search}: We select the $random\_seed$ same as that used to generate the data. Additionally, we search in the range $[-10,\cdots,5]$. We report the performance of the algorithm for varying number of search parameters $n_S$.

\vspace{0.2cm}
\noindent \underline{Grid Search}: We select the range of search as, $[-10,\cdots,5]$. We report the performance of the algorithm for varying number of search parameters $n_S$.

\vspace{0.2cm}
\noindent \underline{HyperOpt}: We select the $random\_seed$ same as that used to generate the data. We select the range of search as, $[-10,\cdots,5]$ and the engine = `tpe'. We report the algorithm's performance for varying values of $n_S$.

\vspace{0.2cm}
\noindent \underline{Spearmint}: We select the range of search as, $[-10,\cdots,5]$ , engine = `GPEIOptChooser' and $grid\_size = 20000$. We report the algorithm's performance for varying values of $n_S$.

\vspace{0.2cm}
\noindent The parameters specific to the problems and the respective data-sets are provided below.

\subsection{Regression using Cookie Data}
\subsubsection{Experiment Parameters} \label{sec_params_cookie}

\underline{Stochastic Hypernetwork Optimization (SHO)}: We keep the following parameter fixed (to default values), $\beta = 0.01$, noise variance $\hat{\lambda} \sim \mathcal{N(\lambda,\sigma)}, \sigma = 10^{-4}$. Changing these values for our experiments did not result in significant improvements. We report the performance of SHO for varying values of $\alpha$. The code is publicly available at: \url{https://github.com/lorraine2/hypernet-hypertraining}

\vspace{0.2cm}
\noindent \underline{Moreau Yosida Regularized HPO (MY-HPO)}: We keep the following parameter fixed, $\rho = 1.0$ and report the results for varying $\alpha, \beta, \delta$. It is well-known in ADMM literature that the selection of $\rho - $ parameter greatly impacts the convergence behavior. Such analyses will be explored in a longer version of the paper.
\vspace{0.2cm}

\noindent Further we fix $n_T$ (i.e. \textsc{MaxIters} in Algorithm \ref{algMYHPO}) $= 2500$, and step size for training the model used for black-box approaches as $\alpha_{train} = 0.001$.

\subsubsection{Additional Results}
The complete set of results with all the parameter settings are provided in Table \ref{tab_results_cookie}. As seen from the results increasing the budget (number of gradient computations $n_g$) to almost $5\times$ that of the bi-level counterparts allows the black-box approaches achieve similar performance.
\begin{table}[h]
\centering
\caption{Comparison between different HPO algorithms for regression problem using Cookie data.} 
\label{tab_results_cookie}
\begin{footnotesize}

\begin{sc}
\begin{tabular}{|l|cccc|}  
\hline 
Method &\specialcell{Train Loss\\($\times 10^-2$)} & \specialcell{Validation Loss\\ ($\times 10^-2$)} & \specialcell{Test Loss\\($\times 10^-2$)} & \specialcell{$n_g$ \\ ($n_T, n_S$)} \\
\hline \hline 
SHO ($\alpha = 5\times 10^{-2}$) & $125.18 \pm 17.08$ & $157.99 \pm 33.43$ & $143.32 \pm 40.32$ & \specialcell{$5000$ \\ ($2500, 2$)}\\
SHO ($\alpha = 1\times 10^{-2}$) & $86.7 \pm 56.08$ & $80.76 \pm 62.36$ & $85.97 \pm 88.78$ & \specialcell{$5000$ \\ ($2500, 2$)} \\
SHO ($\alpha = 5\times 10^{-3}$) & $71.6 \pm 66.96$ & $63.18 \pm 72.85$ & $75.31 \pm 98.78$ & \specialcell{$5000$ \\ ($2500, 2$)}\\[0.5mm] \hline
MyHPO (C) ($\alpha = 0.005, \beta = 0.01, \delta= 0.1$) & $22.9 \pm 15.51$ &  $28.27 \pm 19.94$ & $30.6 \pm 29.05$ & \specialcell{$5000$ \\ ($2500, 2$)}\\
MyHPO (C) ($\alpha = 0.005, \beta = 0.01, \delta= 0.5$) & $7.8 \pm 2.68$ & $17.64 \pm 13.41$ & $17.42 \pm 15.82$ &\specialcell{$5000$ \\ ($2500, 2$)} \\
MyHPO (BT) ($\alpha = 0.005, \beta = 0.01, \delta= 0.5$) & $7.8 \pm 2.68$ & $17.64 \pm 13.41$ & $17.42 \pm 15.82$ & \specialcell{$5000$ \\ ($2500, 2$)}\\
MyHPO (BT) ($\alpha = 0.01, \beta = 0.01, \delta= 0.5$) & $6.85 \pm 5.68$ & $9.74 \pm 6.6$ & $11.31 \pm 13.54$ &\specialcell{$5000$ \\ ($2500, 2$)}\\
MyHPO (BT) ($\alpha = 0.1, \beta = 0.1, \delta= 0.5$) & $5.41 \pm 2.53$ & $6.28 \pm 4.82$ & $6.91 \pm 7.25$ & \specialcell{$5000$ \\ ($2500, 2$)} \\[0.5mm] \hline
Random ($n_S = 2$) & $89.02 \pm 10.36$  & $93.44 \pm 26.4$ & $85.69 \pm 28.97$ & \specialcell{$5000$ \\ ($2500, 2$)} \\
\quad \quad \quad \quad  ($n_S = 10$) & $22.7 \pm 4.18$  & $22.56 \pm 15.03$ & $21.96 \pm 17.05$ & \specialcell{$25000$ \\ ($2500, 10$)} \\ \hline
Grid \quad \quad ($n_S = 2$) & $6.7 \pm 2.21$  & $16.71 \pm 12.91$ & $16.15 \pm 14.76$ & \specialcell{$5000$ \\ ($2500, 2$)}  \\
\quad \quad \quad \quad($n_S = 10$) & $6.7 \pm 2.21$  & $16.71 \pm 12.91$ & $16.15 \pm 14.76$ & \specialcell{$25000$ \\ ($2500, 10$)}\\ \hline
HyperOpt ($n_S = 2$) &$58.11 \pm 7.97$  & $49.77 \pm 19.57$ & $47.29 \pm 23.54$ &  \specialcell{$5000$ \\ ($2500, 2$)}\\
\quad \quad \quad \quad \quad   ($n_S = 10$) &$6.89 \pm 2.19$  & $16.75 \pm 12.94$ & $16.20 \pm 14.79$ &  \specialcell{$25000$ \\ ($2500, 2$)} \\ \hline
Spearmint ($n_S = 2$) &$6.70 \pm 2.21$  & $16.71 \pm 12.91$ & $16.15 \pm 14.76$ & \specialcell{$5000$ \\ ($2500, 2$)} \\
\quad \quad \quad \quad \quad   ($n_S = 10$) &$6.70 \pm 2.21$  & $16.71 \pm 12.91$ & $16.15 \pm 14.76$ & \specialcell{$25000$ \\ ($2500, 10$)} \\ \hline
\end{tabular}
\end{sc}
\end{footnotesize}
\end{table}


\begin{figure*}[h]
\centering
\includegraphics[height=5.0cm]{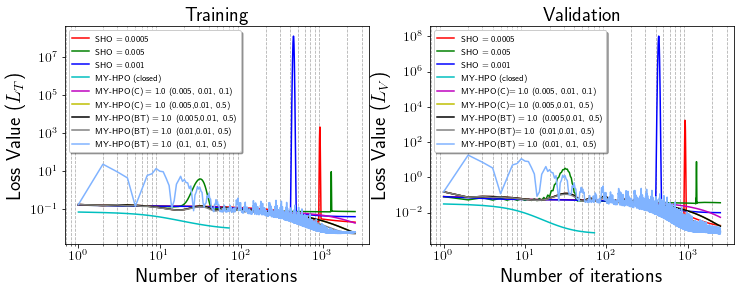}
\caption{Convergence behavior of SHO vs. MY-HPO for different step-sizes for Cookie data using Least Square loss. For MY-HPO we show the $\rho (\alpha,\beta,\delta)$ values.}\label{fig_Cookie_all}
\vspace{-0.2cm}
\end{figure*}


The figure \ref{fig_Cookie_all} illustrate the convergence curve of the algorithms. Fig.  \ref{fig_Cookie_all} illustrates the unstable behavior of SHO as the step size increases. Note that, for larger step sizes the SHO algorithm fails to converge. As the step size gets smaller the algorithm converges; but demonstrate a very slow convergence rate. This is also seen for the other experiments used in this paper. On the contrary MY-HPO can accomodate higher step size and results to better convergence. Additionally, \ref{fig_Cookie_all} also provides a comparison between the MY-HPO algorithm in \ref{algMYHPO} vs. the simplified version of algorithm \ref{algMYHPO_simple}. As seen in the figure (also confirmed in Table \ref{tab_results_closed}), the simplified algorithm provides better per-step iteration cost but incurs much higher overall iterations to convergence.  
   
\newpage

\subsection{Regression using MNIST Data}

\subsubsection{Experiment Parameters}
\underline{Stochastic Hypernetwork Optimization (SHO)}: We fix the parameters same as in \ref{sec_params_cookie}. We vary the $\alpha$ parameter as shown in Table \ref{tab_results_mnist_reg}.

\vspace{0.2cm}
\noindent \underline{Moreau Yosida Regularized HPO (MY-HPO)}: We fix the parameters same as in \ref{sec_params_cookie}. We vary the $\alpha, \beta, \delta$ parameters as shown in Table \ref{tab_results_mnist_reg}.

\vspace{0.2cm}
\noindent Further for this data we fix $n_T$ (i.e. \textsc{MaxIters} in Algorithm \ref{algMYHPO}) $= 3000$, and step size for training the model used for black-box approaches as $\alpha_{train} = 0.001$.

\subsubsection{Additional Results}
\begin{table}[h]
\centering
\caption{Comparison between different HPO algorithms for regression problem using MNIST data.} 
\label{tab_results_mnist_reg}
\begin{footnotesize}

\begin{sc}
\begin{tabular}{|l|cccc|}  
\hline 
Method & \specialcell{Train Loss\\($\times 10^-2$)} & \specialcell{Validation Loss\\ ($\times 10^-2$)} & \specialcell{Test Loss\\($\times 10^-2$)} & \specialcell{$n_g$ \\ ($n_T, n_S$)} \\
\hline \hline 
SHO ($\alpha = 1\times 10^{-2}$) & $32.67 \pm 1.83$ & $32.27 \pm 6.13$ & $32.33 \pm 7.46$ & \specialcell{$6000$ \\ ($3000, 2$)}\\
SHO ($\alpha = 5\times 10^{-3}$) & $27.28 \pm 3.69$ & $25.26 \pm 1.22$ & $24.22 \pm 0.98$ & \specialcell{$6000$ \\ ($3000, 2$)} \\
SHO ($\alpha = 1\times 10^{-3}$) & $26.7 \pm 2.79$ & $24.63 \pm 1.37$ & $23.45 \pm 1.11$ & \specialcell{$6000$ \\ ($3000, 2$)}\\[0.5mm] \hline
MyHPO (C) ($\alpha = 0.001, \beta = 0.001, \delta= 0.001$) & $23.13 \pm 0.66$ &  $23.67 \pm 0.97$ & $22.31 \pm 0.48$ & \specialcell{$6000$ \\ ($3000, 2$)}\\
MyHPO (C) ($\alpha = 0.001, \beta = 0.001, \delta= 0.005$) & $22.63 \pm 0.63$ & $23.69 \pm 1.02$ & $22.32 \pm 0.55$ &\specialcell{$6000$ \\ ($3000, 2$)} \\
MyHPO (BT) ($\alpha = 0.001, \beta = 0.001, \delta= 0.005$) & $22.63 \pm 0.63$ & $23.69 \pm 1.02$ & $22.32 \pm 0.55$ & \specialcell{$6000$ \\ ($3000, 2$)}\\
MyHPO (BT) ($\alpha = 0.01, \beta = 0.01, \delta= 0.05$) & $22.43 \pm 0.61$ & $23.69 \pm 1.11$ & $22.53 \pm 0.72$ &\specialcell{$6000$ \\ ($3000, 2$)}\\
MyHPO (BT) ($\alpha = 0.1, \beta = 0.1, \delta= 0.5$) & $21.75 \pm 0.77$ & $24.16 \pm 1.18$ & $22.85 \pm 0.89$ & \specialcell{$6000$ \\ ($3000, 2$)} \\[0.5mm] \hline
Random ($n_S = 2$) & $22.28 \pm 6.04$  & $25.59 \pm 2.86$ & $24.13 \pm 2.85$ & \specialcell{$6000$ \\ ($3000, 2$)} \\ \hline
Grid \quad \quad ($n_S = 2$) & $17.44 \pm 0.52$  & $26.3 \pm 2.48$ & $24.79 \pm 2.17$ & \specialcell{$6000$ \\ ($3000, 2$)}  \\ \hline
HyperOpt ($n_S = 2$) &$22.84 \pm 5.75$  & $25.02 \pm 2.48$ & $23.85 \pm 2.15$ &  \specialcell{$6000$ \\ ($3000, 2$)}\\ \hline
Spearmint ($n_S = 2$) &$17.44 \pm 0.52$  & $26.30 \pm 2.48$ & $24.79 \pm 2.17$ & \specialcell{$6000$ \\ ($3000, 2$)} \\ \hline
\end{tabular}
\end{sc}
\end{footnotesize}
\end{table}

\begin{figure*}[h]
\centering
\includegraphics[height=5.5cm]{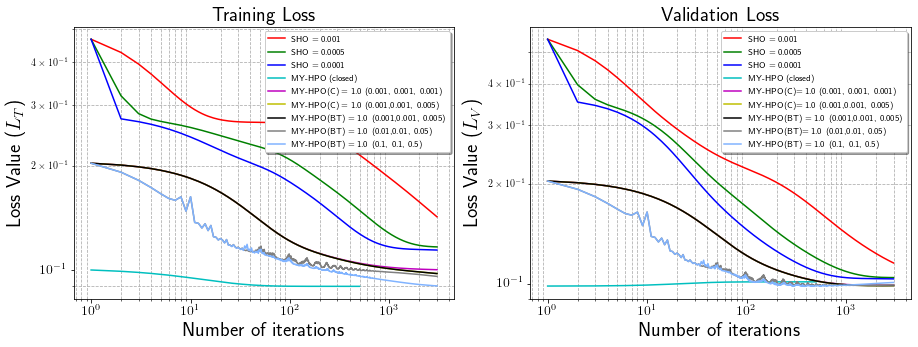}
\caption{Convergence behavior of SHO vs. MY-HPO for different step-sizes for MNIST data using Least Square loss. For MY-HPO we show the $\rho (\alpha,\beta,\delta)$ values.}\label{fig_MNIST_reg_all}
\vspace{-0.2cm}
\end{figure*}

Note that, here further increasing the computation budget (i.e. $n_S = 10$) does not provide additional improvement for the black-box approaches. The results provide similar conclusions,
\begin{enumerate}[nosep]
\item MY-HPO accomodates for higher step sizes and results to better convergence.
\item Back-tracking further ensures descent direction (in each iteration) and leads to better convergence.
\item Increasing the budget (number of gradient computations $n_g$) to almost $5\times$ that of the bi-level counterparts allows the black-box approaches achieve similar performance.
\end{enumerate}

\subsection{Classification using MNIST Data}
\subsubsection{Experiment Parameters}
\underline{Stochastic Hypernetwork Optimization (SHO)}: We fix the parameters same as in \ref{sec_params_cookie}. We vary the $\alpha$ parameter as shown in Table \ref{tab_results_mnist}.

\vspace{0.2cm}
\noindent \underline{Moreau Yosida Regularized HPO (MY-HPO)}: We fix the parameters same as in \ref{sec_params_cookie}. We vary the $\alpha, \beta, \delta$ parameters as shown in Table \ref{tab_results_mnist}.

\vspace{0.2cm}
\noindent Further for this data we fix $n_T$ (i.e. \textsc{MaxIters} in Algorithm \ref{algMYHPO}) $= 500$, and step size for training the model used for black-box approaches as $\alpha_{train} = 0.5$.

\subsubsection{Additional Results}
\begin{table}[h]
\centering
\caption{Comparison between different HPO algorithms for classification problem using MNIST data.} 
\label{tab_results_mnist}
\begin{footnotesize}

\begin{sc}
\begin{tabular}{|l|cccc|}  
\hline 
Method & \specialcell{Train Loss\\($\times 10^-2$)} & \specialcell{Validation Loss\\ ($\times 10^-2$)} & \specialcell{Test Loss\\($\times 10^-2$)} & \specialcell{$n_g$ \\ ($n_T, n_S$)} \\
\hline \hline 
SHO ($\alpha = 5\times 10^{-2}$) & $47.46 \pm 74.13$ & $31.19 \pm 71.85$ & $31.56 \pm 72.45$ & \specialcell{$1000$ \\ ($500, 2$)}\\
SHO ($\alpha = 1\times 10^{-2}$) & $5.58 \pm 2.14$ & $4.98 \pm 1.96$ & $5.32 \pm 1.6$ & \specialcell{$1000$ \\ ($500, 2$)} \\
SHO ($\alpha = 1\times 10^{-3}$) & $6.58 \pm 1.93$ & $5.59 \pm 1.75$ & $5.68 \pm 1.80$ & \specialcell{$1000$ \\ ($500, 2$)}\\[0.5mm] \hline
MyHPO (C) ($\alpha = 0.05, \beta = 0.1, \delta= 0.1$) & $5.17 \pm 0.97$ &  $4.91 \pm 1.65$ & $5.23 \pm 1.49$ & \specialcell{$1000$ \\ ($500, 2$)}\\
MyHPO (C) ($\alpha = 0.05, \beta = 0.1, \delta= 0.5$) & $2.99 \pm 1.57$ & $4.42 \pm 1.89$ & $5.06 \pm 1.83$ &\specialcell{$1000$ \\ ($500, 2$)} \\
MyHPO (BT) ($\alpha = 0.05, \beta = 0.1, \delta= 0.5$) & $2.94 \pm 1.60$ & $4.40 \pm 1.89$ & $5.06 \pm 1.84$ & \specialcell{$1000$ \\ ($500, 2$)}\\
MyHPO (BT) ($\alpha = 0.1, \beta = 0.1, \delta= 0.5$) & $3.17 \pm 1.47$ & $4.40 \pm 1.88$ & $5.02 \pm 1.72$ &\specialcell{$1000$ \\ ($500, 2$)}\\
MyHPO (BT) ($\alpha = 0.1, \beta = 0.5, \delta= 0.75$) & $2.87 \pm 1.68$ & $4.35 \pm 1.94$ & $5.06 \pm 1.85$ & \specialcell{$1000$ \\ ($500, 2$)} \\[0.5mm] \hline
Random ($n_S = 2$) & $35.733 \pm 71.443$  & $19.07 \pm 37.81$ & $19.49 \pm 38.62$ & \specialcell{$1000$ \\ ($500, 2$)} \\
\quad \quad \quad \quad  ($n_S = 25$) & $2.17 \pm 0.93$  & $4.46 \pm 2.03$ & $5.05 \pm 1.89$ & \specialcell{$12500$ \\ ($500, 25$)} \\ \hline
Grid \quad \quad ($n_S = 2$) & $0.3 \pm 0.1$  & $5.31 \pm 2.71$ & $6.22 \pm 2.82$ & \specialcell{$1000$ \\ ($500, 2$)}  \\
\quad \quad \quad \quad($n_S = 25$) & $2.53 \pm 1.23$  & $4.37 \pm 1.88$ & $5.05 \pm 1.84$ & \specialcell{$12500$ \\ ($500, 25$)}\\ \hline
HyperOpt ($n_S = 2$) &$5.18 \pm 0.35$  & $4.91 \pm 1.46$ & $5.19 \pm 1.44$ &  \specialcell{$1000$ \\ ($500, 2$)}\\
\quad \quad \quad \quad \quad   ($n_S = 25$) &$2.44 \pm 1.43$  & $4.40 \pm 1.88$ & $4.18 \pm 1.97$ &  \specialcell{$12500$ \\ ($500, 25$)} \\ \hline
Spearmint ($n_S = 2$) &$0.30 \pm 0.10$  & $5.31 \pm 2.71$ & $6.22 \pm 2.82$ & \specialcell{$1000$ \\ ($500, 2$)} \\
\quad \quad \quad \quad \quad   ($n_S = 25$) &$1.89 \pm 1.55$  & $4.63 \pm 2.19$ & $5.36 \pm 2.27$ & \specialcell{$12500$ \\ ($500, 25$)} \\ \hline
\end{tabular}
\end{sc}
\end{footnotesize}
\end{table}

\begin{figure*}[h]
\centering
\includegraphics[height=5.5cm]{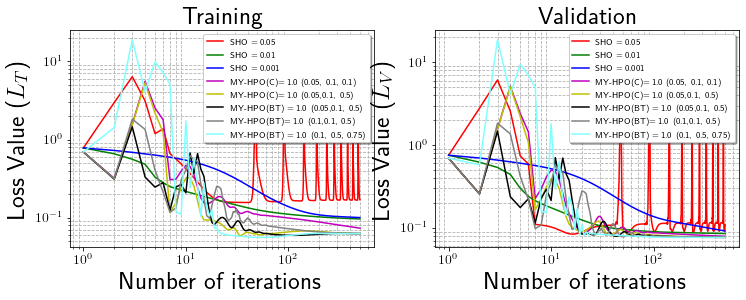}
\caption{Convergence behavior of SHO vs. MY-HPO for different step-sizes for MNIST data using Logistic loss. For MY-HPO we show the $\rho (\alpha,\beta,\delta)$ values.}\label{fig_MNIST_class_all}
\vspace{-0.2cm}
\end{figure*}

\newpage

\subsection{Classification using GTSRB Data}
\subsubsection{Experiment Parameters}
\underline{Stochastic Hypernetwork Optimization (SHO)}: We fix the parameters same as in \ref{sec_params_cookie}. We vary the $\alpha$ parameter as shown in Table \ref{tab_results_mnist}.

\vspace{0.2cm}
\noindent \underline{Moreau Yosida Regularized HPO (MY-HPO)}: We fix the parameters same as in \ref{sec_params_cookie}. We vary the $\alpha, \beta, \delta$ parameters as shown in Table \ref{tab_results_mnist}.

\vspace{0.2cm}
\noindent Further for this data we fix $n_T$ (i.e. \textsc{MaxIters} in Algorithm \ref{algMYHPO}) $= 500$, and step size for training the model used for black-box approaches as $\alpha_{train} = 0.5$.
\subsubsection{Additional Results}
\begin{table}[h]
\centering
\caption{Comparison between different HPO algorithms for classification problem using GTSRB data.} 
\label{tab_results_gtsrb}
\begin{footnotesize}
\begin{sc}
\begin{tabular}{|l|cccc|}  
\hline 
Method & \specialcell{Train Loss\\($\times 10^-2$)} & \specialcell{Validation Loss\\ ($\times 10^-2$)} & \specialcell{Test Loss\\($\times 10^-2$)} & \specialcell{$n_g$ \\ ($n_T, n_S$)} \\
\hline \hline 
SHO ($\alpha = 5\times 10^{-2}$) & $27.48 \pm 20.54$ & $25.38 \pm 19.09$ & $25.61 \pm 20.27$ & \specialcell{$1000$ \\ ($500, 2$)}\\
SHO ($\alpha = 1\times 10^{-2}$) & $6.83 \pm 1.04$ & $14.35 \pm 2.34$ & $14.04 \pm 1.82$ & \specialcell{$1000$ \\ ($500, 2$)} \\
SHO ($\alpha = 5\times 10^{-3}$) & $6.89 \pm 1.13$ & $14.44 \pm 2.41$ & $14.14 \pm 1.85$ & \specialcell{$1000$ \\ ($500, 2$)}\\[0.5mm] \hline
MyHPO (C) ($\alpha = 0.05, \beta = 0.1, \delta= 0.1$) & $7.98 \pm 0.74$ &  $14.81 \pm 2.2$ & $14.53 \pm 1.61$ & \specialcell{$1000$ \\ ($500, 2$)}\\
MyHPO (C) ($\alpha = 0.05, \beta = 0.1, \delta= 0.75$) & $3.26 \pm 2.50$ & $14.14 \pm 2.38$ & $13.80 \pm 2.25$ &\specialcell{$1000$ \\ ($500, 2$)} \\
MyHPO (BT) ($\alpha = 0.05, \beta = 0.1, \delta= 0.75$) & $3.26 \pm 2.50$ & $14.14 \pm 2.38$ & $13.80 \pm 2.25$ &\specialcell{$1000$ \\ ($500, 2$)}\\
MyHPO (BT) ($\alpha = 0.1, \beta = 0.5, \delta= 0.75$) & $4.37 \pm 2.11$ & $14.09 \pm 2.36$ & $13.77 \pm 2.11$ &\specialcell{$1000$ \\ ($500, 2$)}\\
MyHPO (BT) ($\alpha = 0.5, \beta = 0.5, \delta= 0.75$) & $4.31 \pm 1.22$ & $13.98 \pm 2.33$ & $13.59 \pm 1.98$ & \specialcell{$1000$ \\ ($500, 2$)} \\[0.5mm] \hline
Random ($n_S = 2$) & $47.48 \pm 117.38$  & $39.85 \pm 75.38$ & $38.60 \pm 70.63$ & \specialcell{$1000$ \\ ($500, 2$)} \\
\quad \quad \quad \quad  ($n_S = 10$) & $4.01 \pm 2.07$  & $14.19 \pm 2.24$ & $13.63 \pm 1.91$ & \specialcell{$5000$ \\ ($500, 10$)} \\ \hline
Grid \quad \quad ($n_S = 2$) & $0.09 \pm 0.01$  & $23.55 \pm 6.60$ & $21.81 \pm 5.23$ & \specialcell{$1000$ \\ ($500, 2$)}  \\
\quad \quad \quad \quad($n_S = 10$) & $3.53 \pm 0.49$  & $13.87 \pm 2.34$ & $13.40 \pm 2.03$ & \specialcell{$5000$ \\ ($500, 10$)}\\ \hline
HyperOpt ($n_S = 2$) &$11.91 \pm 30.50$  & $18.9 \pm 11.7$ & $17.62 \pm 8.52$ &  \specialcell{$1000$ \\ ($500, 2$)}\\
\quad \quad \quad \quad \quad   ($n_S = 10$) &$3.3 \pm 1.46$  & $13.94 \pm 2.40$ & $13.58 \pm 2.17$ &  \specialcell{$5000$ \\ ($500, 10$)} \\ \hline
Spearmint ($n_S = 2$) &$0.09 \pm 0.01$  & $23.55 \pm 6.60$ & $21.81 \pm 5.23$ & \specialcell{$1000$ \\ ($500, 2$)} \\
\quad \quad \quad \quad \quad   ($n_S = 10$) &$2.81 \pm 5.63$  & $19.73 \pm 6.28$ & $19.20 \pm 5.59$ & \specialcell{$5000$ \\ ($500, 10$)} \\ \hline
\end{tabular}
\end{sc}
\end{footnotesize}
\end{table}

\end{document}